# Design a Technology Based on the Fusion of Genetic Algorithm, Neural network and Fuzzy logic

**Raid R. Al-Nima**        **Fawaz S. Abdullah**        **Ali N. Hamoodi**
*Technical Engineering College /*
*Northern Technical University /*
*Mosul / Iraq*

**ABSTRACT**

This paper describes the design and development of a prototype technique for artificial intelligence based on the fusion of genetic algorithm, neural network and fuzzy logic. It starts by establishing a relationship between the neural network and fuzzy logic. Then, it combines the genetic algorithm with them. Information fusions are at the confidence level, where matching scores can be reported and discussed. The technique is called the Genetic Neuro-Fuzzy (GNF). It can be used for high accuracy real-time environments.

## 1. INTRODUCTION

Many authors focused on Fuzzy logic [1], Artificial Neural Network (ANN) [2] and Genetic Algorithm (GA) [3]. Then other techniques adopt a fusion between two of them, like a neuro-fuzzy research for speaker recognition [4] and a genetic neural network for signature recognition [5]. Also there are many researches about their fusion for an application, like using five layer neural network with learning algorithm-genetic to determine the optimal collision-free path for Obstacle Avoidance of a wheeled mobile robot [6]. Moreover, there is a research about comparative analysis of the fuzzy, Neuro-fuzzy and Fuzzy-GA approaches which performed to evaluate the reusability of software components and Fuzzy-GA results outperform the other used approaches [7].

The point of fuzzy logic is to map an input space to an output space, and the primary mechanism for doing this is a list of if-then statements called rules. All rules are evaluated in parallel, and the order of the rules is unimportant. The rules themselves are useful because they refer to variables and the adjectives that describe those variables [8].

ANN is an information-processing system that has certain performance characteristics in common with biological neural networks. Artificial neural networks have been developed as generalizations of mathematical models of human cognition or neural biology [2].

The GA is a method for solving both constrained and unconstrained optimization problems that is based on natural selection, the process that drives biological evolution [9].

So, among various combinations of methodologies in soft computing, the one that has highest visibility at juncture is that of fuzzy logic and neurocomputing, leading to so-called neuro-fuzzy systems [8].

To apply GAs to neural networks. Some aspects that can be evolved are the weights in a fixed network, the network architecture (i.e., the number of units and their interconnections can change), and the learning rule used by the network [3].

The aim of this research is to apply a new combination technique named Genetic Neuro-Fuzzy (GNF) network.

## 2. Fuzzy Logic:

Fuzzy logic starts with the concept of a fuzzy set. A fuzzy set is a set without a crisp, clearly defined boundary. It can contain elements with only a partial degree of membership.

A membership function must really satisfy is that it must vary between 0 and 1. The function itself can be an arbitrary curve whose shape can be defined as a function that suits from the point of view of simplicity, convenience, speed, and efficiency.

A classical set might be expressed as:

$$A = \{x \mid x > 6\} \dots\dots\dots\dots\dots (1)$$

A fuzzy set is an extension of a classical set. If $X$ is the universe of discourse and its elements are denoted by $x$, then a fuzzy set $A$ in $X$ is defined as a set of ordered pairs.

$$A = \{x, \ \mu A(x) \mid x \in X\} \dots\dots\dots\dots(2)$$

$\mu A(x)$ is called the membership function (or MF) of $x$ in $A$. The membership function maps each element of $X$ to a membership value between 0 and 1.

Moreover to resolve the statement A AND B, where A and B are limited to the range (0,1), by using the function min(A,B). Using the same reasoning, the OR operation can be replaced with the max function, so that A OR B becomes equivalent to max(A,B). Finally, the operation NOT A becomes equivalent to the operation 1-A.

The intersection of two fuzzy sets $A$ and $B$ is specified in general by a binary mapping $T$, which aggregates two membership functions as follows:

$$\mu A \cap B(x) = T(\mu A(x), \ \mu B(x)) \quad \dots\dots(3)$$

Like fuzzy intersection, the fuzzy union operator is specified in general by a binary mapping $S$:

$$\mu A \cap B(x) = S(\mu A(x), \ \mu B(x)) \quad \dots\dots(4)$$

The input for the defuzzification process is a fuzzy set (the aggregate output fuzzy set) and the output is a single number. [8]

## 3. Neural Network:

The transmission of signals in biological neurons through synapses is a complicated chemical process in which specific transmitter substances are released from the sending side of the synapse. The effect is to raise or lower the electrical potential inside the body of the receiving cell. The neuron fires if the potential reaches a threshold. This neuron model is widely used in ANN with some variations. Fig. 1 shows the architecture of simple neuron model.

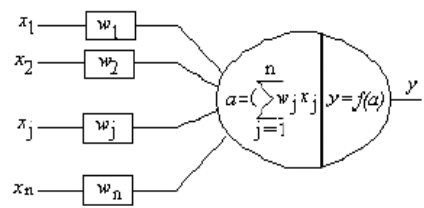

**Figure 1: Simple artificial neuron model**

The artificial neuron presented in the above figure has n inputs, denoted as $x_1, x_2, \dots, x_n$. Each line connecting these inputs to the neuron is assigned a weight, denoted as $w_1, w_2, \dots, w_n$, respectively.

The action, which determines whether the neuron is to be fired or not, is given by the formula:

$$a = \sum_{j=1}^{n} w_j x_j \qquad \dots\dots\dots\dots\dots (5)$$

The output of the neuron is a function of its action:

$$y = f(a) \qquad \dots\dots\dots\dots\dots\dots (6)$$

Originally the neuron output function *f(a)* was a threshold function. However, linear, ramp and sigmoid functions are also widely used today. An ANN system consists of a number of artificial neurons and a huge number of interconnections among them. According to the structure of the connections [4,10].

## 4. Genetic Algorithm (GA):

The GA repeatedly modifies a population of individual solutions. At each step, the GA selects individuals at random from the current population to be parents and uses them to produce the children for the next generation. Over successive generations, the population "evolves" toward an optimal solution [9].

The following outline summarizes how the GA works:
1. The algorithm begins by creating a random initial population.
2. The algorithm then creates a sequence of new populations. At each step, the algorithm uses the individuals in the current generation to create the next population. To create the new population, the algorithm performs the following steps:
   a. Scores each member of the current population by computing its fitness value.
   b. Scales the raw fitness scores to convert them into a more usable range of values.
   c. Selects members, called parents, based on their fitness.
   d. Some of the individuals in the current population that have lower fitness are chosen as elite. These elite individuals are passed to the next population.
   e. Produces children from the parents. Children are produced either by making random changes to a single parent—mutation—or by combining the vector entries of a pair of parents—crossover.
   f. Replaces the current population with the children to form the next generation.
3. The algorithm stops when one of the stopping criteria is met. [9]

## 5. GNF suggested:

The idea of GNF network suggested are given from the adaptive neuro-fuzzy inference system idea.

The acronym ANFIS derives its name from adaptive neuro-fuzzy inference system. The basic idea behind these neuro-adaptive learning techniques is very simple. These techniques provide a method for the fuzzy modeling procedure to learn information about a data set, in order to compute the membership function parameters that best allow the associated fuzzy inference system to track the given input/output data. This learning method works similarly to that of neural networks [8].

Using a given input/output data set, the constructed fuzzy inference system (FIS) has membership function parameters tuned (adjusted) using either a backpropagation algorithm alone, or in combination with a least squares type of method. This allows fuzzy systems to learn from the data they are modeling [8].

Moreover, the activation function in neural network has spread through a feedforward network, the resulting activation pattern on the output units encodes the network's "answer" to the input. In most applications, the network learns a correct mapping between input and output patterns via a learning algorithm [3]. Then to apply the genetic algorithm evolving the weights in a fixed network can be used to mapping between input and output patterns. The fitness which can be used is shown in equation (5) below:

$$\delta_k = \sum_{k=1}^{m} |t_k - y_k| \qquad \ldots\ldots\ldots\ldots ..(7)$$

Where each output unit in neural network ($y_k$, $k = 1\ldots\ldots.m$) receives a target pattern ($t_k$) corresponding to the input training pattern to computes its error information term ($\delta_k$).

## 5.1 Fuzzy logic example:

We'll starting with a basic description of a two-input, one-output tipping problem, See Fig. 2. The Basic tipping problem given a number between 0 and 10 that represents the quality of service at a restaurant (where 10 is excellent), and another number between 0 and 10 that represents the quality of the food at that restaurant (again, 10 is excellent). The starting point is to write down the three golden rules of tipping, based on years of personal experience in restaurants:

1. If the service is poor or the food is rancid, then tip is cheap.
2. If the service is good, then tip is average.
3. If the service is excellent or the food is delicious, then tip is generous.

We'll assuming that an average tip is 15%, a generous tip is 25%, and a cheap tip is 5%. [8]

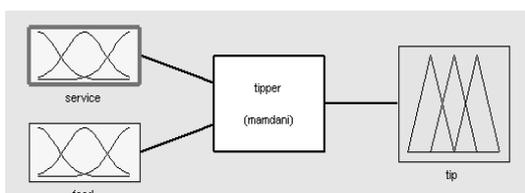

**Figure 2: Fuzzy system for tipper problem**

The membership functions which used the two inputs and the output of this system is shown in Fig. 3.

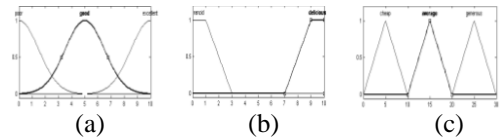

**Figure 3: Membership functions for inputs and output:**
**a) 1st Input variable "Service"**
**b) 2nd Input variable "Food"**
**c) Output variable "Tip"**

## 5.2 Applying Nero-fuzzy method:

Now, to apply ANN in the fuzzy tipper system using a given input/output data set whose membership function parameters are adjusted using a backpropagation algorithm. This allows the tipper fuzzy systems to learn from the data they are modeling. Fig. 4 shows the architecture of suggested backpropagation neural network.

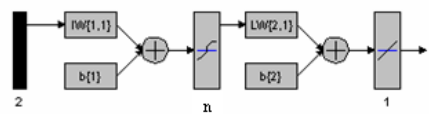

**Figure 4: Suggested backpropagation network architecture**

The Backpropagation network which is suggested has two nodes in the input layer, *n* nodes in the hidden layer (in this example taken 50 nodes) and one node in the output layer. The activation functions used are tan-sigmoid activation functions in the hidden layer and pure-linear activation function in the output layer. By this topology the Backpropagation network is able to get the output fuzzy sample (tipper value) from the two input fuzzy samples (service and food).

## 5.3 Applying GA to the neuro-fuzzy:

Moreover, GA can be implemented in a neural network weights. E.g. under the GA are equivalent to one backpropagation iteration, since backpropagation on a given training consists of two parts: the forward propagation of activation (and the calculation of errors at the output units) and the backward error propagation (and adjusting of the weights). The GA performs only the first part. Since the second part requires more computation, two GA evaluations takes less than half the computation of a single backpropagation iteration. Fig. 5 shows the crossover and mutation processes examples in genetic neural network.

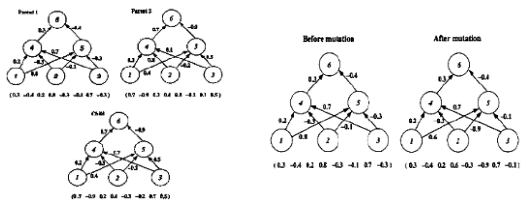

(a)                    (b)

**Figure 5: Genetic neural network crossover and mutation:**
**a) Crossover process example**
**b) Mutation process example**

## 6. The results

The Rule Viewer (for the fuzzy tipper example) displays a roadmap of the fuzzy inference process, See Fig. 6. A single figure window can be seen with 10 small plots nested in it. The three small plots across the top of the figure represent the antecedent and consequent of the first rule. Each rule is a row of plots, and each column is a variable. The first two columns of plots (the six input plots) show the membership functions referenced by the antecedent, or the if-part of each rule. The third column of plots (the three blue plots) shows the membership functions referenced by the consequent, or the then-part of each rule. Notice that under food, there is a plot which is blank. This corresponds to the characterization of none for the variable food in the second rule. The fourth plot in the third column of plots represents the aggregate weighted decision for the given inference system. This decision will depend on the input values for the system.

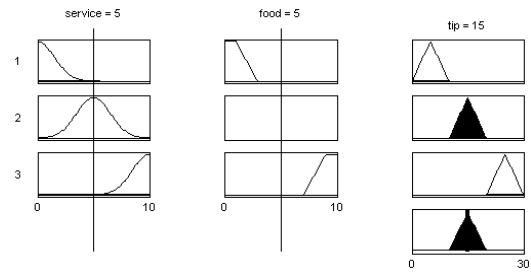

**Figure 6: Output fuzzy viewer rules**

The main problem of fuzzy system is not real-time in execution especially for multiple values of inputs. Therefore, the neuro-fuzzy implemented by backpropagation neural network which trained for all fuzzy tipper probability. Fig. 7 shows the neural network training curve which proved that the backpropagation network reached to its goals. The training stopped by the tolerance of error equal to 0.001.

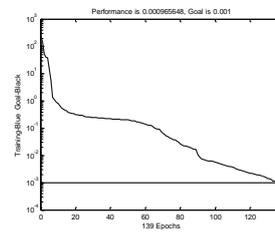

**Figure 7: Backpropagation neural network training**

Then GA can be applied on the neuro-fuzzy tipper example in order to improve its results. Equation (7) can be used as a fitness function, then its error will be compared with the tolerance equal to 0.00001. Fig. 8 shows the results of output errors.

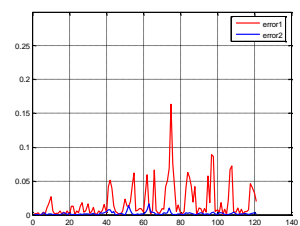

**Figure 8: The differences of errors between the neuro-fuzzy (error1) and suggested GNF (error2)**

From the previous figure, the first error represent the neuro-fuzzy output error values. The second error represents the suggested GNF output error values, it has the best error closed to the x-axis compared with the first error which is so far.

The GNF suggested technique consists from the three main parts: Fuzzy, neural and genetic. Also, it represents the combination of them. See Fig. 9 for suggested GNF block diagram.

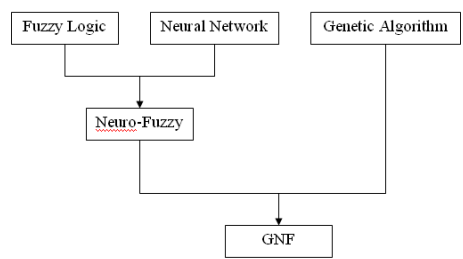

**Figure 9: Suggested GNF block diagram**

So, suggested GNF solve the real-time problem of fuzzy logic for multiple input values by using Neuro-fuzzy method. Also, improve the neural network outputs by using the GA for training.

The GNF can be considered as one of suggestable techniques as in [11-58].

## 7. Conclusions

In this research a new technique is suggested for artificial intelligence named genetic neuro-fuzzy (GNF) network. This technique is base on the fusion among fuzzy logic, neural network and GA. Its idea has taken from the combination between fuzzy and neural network (neuro-fuzzy), also from the combination between neural network and genetic algorithm (genetic neural network). The obtained results may be summarized by the following points:

- Suggested GNF network has more powerful results than neuro-fuzzy and genetic neural network.

- Suggested GNF solve the real-time problem of fuzzy logic for multiple values of inputs because of using neural network to train all probability of fuzzy system values.
- Suggested GNF improve the neural network outputs because of using genetic algorithm to reach the final weights.
- The output error of Neuro-fuzzy compared with tolerance equal to 0.001 while the output error of suggested GNF can be compared with tolerance reach to 0.00001.
- Suggested GNF can be used in high accuracy real-time environments.